\documentclass[conference]{IEEEtran}
\IEEEoverridecommandlockouts

\usepackage{amsmath,amssymb,amsfonts}
\usepackage{algorithmic}
\usepackage{graphicx}
\usepackage{import}
\usepackage{tikz}
\usepackage{cite}
\usepackage{hyperref}
\usepackage{pgf}
\usepackage{xcolor}
\usepackage{listings}
\lstdefinestyle{Bash}
{language=bash,
basicstyle=\small,
keywordstyle=[2]{\bfseries},
morekeywords=[2]{Total,P,S,W},
}

\usepackage{multirow}
\usepackage{textcomp}
\usepackage{xcolor}

\def\BibTeX{{\rm B\kern-.05em{\sc i\kern-.025em b}\kern-.08em
    T\kern-.1667em\lower.7ex\hbox{E}\kern-.125emX}}
\def\layersep{2.5cm}

\begin{document}
\title{Application of Machine Learning to Sleep Stage Classification}
\author{\IEEEauthorblockN{1\textsuperscript{st} Andrew Smith}
\IEEEauthorblockA{\textit{Department of Computer Science and Engineering} \\
\textit{(University of South Carolina)}\\
Columbia, SC 29208 USA  \\
aks3@email.sc.edu}
\and
\IEEEauthorblockN{2\textsuperscript{nd} Hardik Anand}
\IEEEauthorblockA{\textit{Department of Computer Science and Engineering} \\
\textit{(University of South Carolina)}\\
Columbia, SC 29208 USA  \\
hanand@email.sc.edu}
\and
\IEEEauthorblockN{3\textsuperscript{rd} Snezana Milosavljevic}
\IEEEauthorblockA{\textit{Department of Pharmacology, Physiology, and Neuroscience} \\
\textit{University of South Carolina School of Medicine}\\
Columbia, SC 29208 USA \\
snezana.milosavljevic@uscmed.sc.edu}
\and
\IEEEauthorblockN{4\textsuperscript{th} Katherine M. Rentschler}
\IEEEauthorblockA{\textit{Department of Pharmacology, Physiology, and Neuroscience} \\
\textit{University of South Carolina School of Medicine}\\
Columbia, SC 29208 USA \\
katherine.rentschler@uscmed.sc.edu}
\and
\IEEEauthorblockN{5\textsuperscript{th} Ana Pocivavsek}
\IEEEauthorblockA{\textit{Department of Pharmacology, Physiology, and Neuroscience} \\
\textit{University of South Carolina School of Medicine}\\
Columbia, SC 29208 USA \\
ana.pocivavsek@uscmed.sc.edu}
\and
\IEEEauthorblockN{6\textsuperscript{th} Homayoun Valafar}
\IEEEauthorblockA{\textit{Department of Computer Science and Engineering} \\
\textit{(University of South Carolina)}\\
Columbia, SC 29208 USA  \\
homayoun@cse.sc.edu}
}
\maketitle
\begin{abstract}
Sleep studies are imperative to recapitulate phenotypes associated with sleep loss and uncover mechanisms contributing to psychopathology. Most often, investigators manually classify the polysomnography into vigilance states, which is time-consuming, requires extensive training, and is prone to inter-scorer variability. While many works have successfully developed automated vigilance state classifiers based on multiple EEG channels, we aim to produce an automated and open-access classifier that can reliably predict vigilance state based on a single cortical electroencephalogram (EEG) from rodents to minimize the disadvantages that accompany tethering small animals via wires to computer programs. Approximately 427 hours of continuously monitored EEG, electromyogram (EMG), and activity were labeled by a domain expert out of 571 hours of total data. Here we evaluate the performance of various machine learning techniques on classifying 10-second epochs into one of three discrete classes: paradoxical, slow-wave, or wake. Our investigations include Decision Trees, Random Forests, Naive Bayes Classifiers, Logistic Regression Classifiers, and Artificial Neural Networks. These methodologies have achieved accuracies ranging from approximately 74\% to approximately 96\%. Most notably, the Random Forest and the ANN achieved remarkable accuracies of 95.78\% and 93.31\%, respectively. Here we have shown the potential of various machine learning classifiers to automatically, accurately, and reliably classify vigilance states based on a single EEG reading and a single EMG reading.
\end{abstract}
\section{Introduction}
Nearly 70 million Americans are afflicted by chronic sleep disorders or intermittent sleep disturbances that negatively impact health and substantially burden our health system. Sleep is essential for optimal health. Sleep is one of the most critical and ubiquitous biological processes, next to eating and drinking.  It has been shown that there is no clear evidence of the existence of an animal species that does not sleep \cite{a1}. Sleep constitutes about 30\% of the human lifespan. Assessment of sleep quality is multifactorial and is composed of adequate duration, good quality, appropriate timing and regularity, and the absence of sleep disturbances or disorders. 
Sleep duration is used as a metric to describe the standard of healthy sleep. The American Academy of Sleep Medicine (AASM) and Sleep Research Society (SRS) issued a consensus statement recommending ``adults should sleep 7 or more hours per night on a regular basis to promote optimal health"\cite{a2}. However, sufficient sleep is severely undervalued as a necessary biological process for maintaining proper mental health. A recent survey from the Center for Disease Control and Prevention (CDC) found that only 65\% of adults reported a healthy duration of sleep \cite{a3}.

From a translational perspective, animal studies are imperative to recapitulate phenotypes associated with sleep loss (hyperarousal, cognitive impairment, slowed psychomotor vigilance, behavioral despair) and uncover mechanisms contributing to psychopathology, with the added benefit of homogeneity within rodent subjects \cite{b1,b2}. Sleep studies are readily conducted in small animals by implanting electrodes to obtain electroencephalogram (EEG) and electromyogram (EMG). Sleep is categorized into two major classes, non-rapid eye movement (NREM) and rapid eye movement (REM) sleep, and arousal is classified as wake. Most often, investigators manually classify the polysomnography into vigilance states, and this practice is time-consuming and also greatly limits the size of a study’s data set. To accurately classify vigilance states, investigators undergo extensive training, yet the subjective nature of classifying limits inter-scorer reliability.

Several automated vigilance state classifiers have been established, and nearly all of these algorithms rely on multi-channel EEG data and local field potential (LFP) signaling oscillations within the brain \cite{b3,b4,b5,b6,b7,b8,b9}. The advantages of multi-channel systems are outweighed by the disadvantage of tethering small animals to transmit signals via wired connections to computer programs. Tethered animals are combating confounds including limited mobility within recording cages and potential impacts on natural sleep states \cite{b10,b11,b12}. For these reasons, it is advantageous to automate vigilance state classification with telemetric battery devices that are surgically implanted to open a single EEG and EMG from each small animal. Consistent with the principles of Information Theory, data collected from a single EEG channel significantly increases the complexity of sleep-state identification for humans and automated approaches. Therefore, the goal of this research has been to produce an open access and automated classifier that can reliably predict vigilance state based on a single cortical EEG from rodents. To that end, we have evaluated several of the commonly used Machine Learning techniques for suitability and success in this task. Our investigations include Decision Trees, Naive Bayes Classifiers, Random Forests, and Artificial Neural Networks.  An Artificial Neural Network (ANN) has been developed to examine and ascertain the sleep state of each animal.
\section{Methodology}
\subsection{Sleep EEG/EMG Data Collection}
Adult Wistar rats (n=8) were used in experiments in a facility fully accredited by the American Association for the Accreditation of Laboratory Animal Care. Animals were kept on a 12/12 h light-dark cycle. All protocols were approved by the Institutional Animal Care and Use Committee at the University of South Carolina and were in accordance with the National Institutes of Health Guide for the Care and Use of Laboratory Animals.  

Rats were implanted with EEG/EMG telemetry devices (PhysioTel HD-S02, Data Science International, St. Paul, MN), as previously described \cite{b12,b13,b14,b15}. Briefly, under isoflurane anesthesia, animals were placed in a stereotaxic frame. The transmitter device was intraperitoneally implanted through a dorsal incision of the abdominal region. After an incision at the midline of the head was made, EEG leads were secured to two surgical screws inserted into 0.5-mm burr holes at 2.0 mm anterior/1.5 mm lateral and 7.0 mm posterior/-1.5 mm lateral to bregma. Two EMG leads were inserted into the dorsal cervical neck muscle about 1.0 mm apart and sutured into place. The skin was sutured, and animals were recovered for a minimum of 7 days prior to experimentation. 
Sleep data were acquired in a quiet, designated room where rats remained undisturbed for the duration of recording using Ponemah 6.10 software (DSI). Digitized signal data were imported into NeuroScore 3.0 (DSI) and powerband frequencies (0 to 20 Hz) in 0.5 Hz increments were exported to CSV formatted flat files. 
\subsection{Data Processing and Annotation}
Approximately 571 hours of continuously monitored electroencephalogram (EEG), electromyogram (EMG), and activity were recorded across the eight laboratory rodents. The collection of data was then partitioned into 10-second \textit{epochs}, or segments, and labeled by a domain expert. Approximately 427 hours of the recorded 571 hours have been manually labeled by a domain expert. In this study, our focus has been based on the manually annotated 427 hours of the data. The remaining 144 hours of unlabelled data will be used in the future for more rigorous testing of our automated classification method. Based on the polysomnogram (PSG), an expert labels each 10-second epoch as one of three discrete classes: Paradoxical, Slow-wave, or Wake. Paradoxical sleep is also known as rapid-eye-movement (REM) sleep, because of the paradox of the high-frequency brain waves mimicking wakefulness despite being asleep. Slow-Wave sleep, characterized by low frequency, high amplitude brain waves, is also known as non-rapid eye movement sleep (NREM), and it constitutes all sleep that is not REM sleep. Finally, the ``wake" classification is given to a PSG that elicits characteristics of wakefulness. These sleep stages constitute all three \textit{classes} and will be referred to as P, S, and W, which correspond to Paradoxical, Slow-wave, and Wake states, respectively. 

42 discrete features were extracted from each 10-second epoch of continuous EEG, EMG, and activity signals. To obtain the first 40 features, the 10-second epoch is transformed from the time domain into the frequency domain using Discrete Fourier Transformation, then partitioned into 40 channels of equal width from 0 to 20 Hz. Thus, the first channel is the EEG from 0 to 0.5 Hz, the second channel is the EEG from 0.5 to 1 Hz, and so on. The 41st feature is that of the EMG, which is averaged over the 10-second epoch. The 42nd and final feature is the ``Activity" feature, which is a derived parameter from \href{https://support.datasci.com/hc/en-us/articles/115005030328}{Ponemah} (indicating the level of an animal's activity) which depends on the transmitter model, the speed with which the transmitter moves, outside radio interference, and variations from sensor to sensor.

\subsection{Input Formalization}
Many tactics in modern machine learning and artificial intelligence have been presented that aim to formalize the use of temporal data in the tasks of prediction and classification. Although the proper formulation of input can have a substantial impact on the trainability and the outcome of ML developments, in this investigation, we have implemented the most prevalent and natural approach. More specifically, the input to our ML activities consisted of a concatenation of five consecutive processed data (42 channels) that summarize a 10-second epoch from the raw continuous data. We postulate that a single epoch (summary of 10 seconds of data) is not the optimal temporal representation of time-series data for use in ML applications. Therefore, we choose to reformat the data such that each sample spans a longer period, theoretically providing each classifier with more temporal information and confidence. Five consecutive epochs encapsulate 50 seconds of the temporal signal, to which we will refer as the network input in the remainder of this report. Before this windowing, the data consisted of 154043 rows and 43 columns. Each row, which itself constitutes a 10-second epoch, consists of the 42 features and 1 output label describing the three stages of vigilance. 

This mechanism of input data creation results in the following class distribution:

\begin{lstlisting}[style=Bash]
Total: 154039
P: 10028 (6.50% of total)
S: 64539 (41.77% of total)
W: 79472 (51.73% of total).
\end{lstlisting} 

This data set was created by a simple moving window, where the first windowed sample will consist of samples 0-4 in the original data. The second windowed sample consisted of samples 1-5 in the original data, and the final windowed sample consisted of samples (n-5)-(n-1) in the original data. Since each input spans five individual vigilance states, various methods of arriving at a single output (given from five outputs) can be envisioned. In this work, we choose to label each windowed sample with the most frequent class in the window. If there is a tie, we label the sample with the label of the 10-second epoch in the center.
The total number of samples after the simple moving windowing has 4 samples less than the original data, which is well known to be the case when windowing data. Thus, each row in the input data consists of 210 features (5 sets of 42 features), and 1 label.

\subsection{Data Balancing}
The class distribution in the raw windowed data is unequal. Ideally, there would be an equal representation of each class, where each class constitutes 33\% of the data, in this instance. However, class P is severely underrepresented at 6.51\% of the total data, or 10028 samples. Classes S and W are over-represented at 41.90\% and 51.59\%, or 64539 samples and 79472 samples, respectively. In balancing the representation between classes, we aim to replicate copies of classes to the data while minimizing total samples. Minimizing total samples is important to improve training speed. By concatenating complete copies of any given class to itself, we aim to ensure that the model adequately generalizes to the entire class sample. Therefore, we concatenate 7 additional copies of the entire P class to the data, which produced the following and improved class distribution:
\begin{lstlisting}[style=Bash]
Total: 224235
P: 80224 (35.78% of total)
S: 64539 (28.78% of total)
W: 79472 (35.44% of total).
\end{lstlisting}

\subsection{Training, Validation, and Testing Split and Shuffle}
To finalize data preparation before training a classifier, the data must be shuffled and partitioned into training, validation, and testing sets. To perform the shuffling and partitioning, we use a function from the popular machine learning module in python \href{https://scikit-learn.org/stable/}{scikit-learn}. This function randomly shuffles the data and partitions it into training and testing data. We split the data into 80\% training and 20\% testing data. The only machine learning classifier which uses a validation set is the Artificial Neural Network. For the Artificial Neural Network, we further split the training data into 80\% training, 20\% validation, which constitutes 64\% and 16\% of the entire data. This process was performed only once to keep the training, testing, and validation (when needed) sets constant across each ML technique. A consistent training/testing set will help to establish a more consistent comparison of performances across multiple techniques while also reducing the data preparation time. 

\subsection{Evaluation of Machine Learning Classifiers}
We propose to evaluate various machine learning techniques to classify vigilance states. Following are brief descriptions of Decision Trees, Random Forests, Naive Bayes Classifiers, Logistic Regression Classifiers, and Artificial Neural Networks. 
We aim to find the simplest model that achieves the highest accuracy.
\subsection{Decision Tree}
A Decision Tree Classifier is a supervised machine learning algorithm that performs classification tasks. The algorithm behind a Decision Tree makes a sequence of decisions based on input features, one decision at each node along a path from the root to a leaf of the tree. The leaf node that the algorithm ends on for any given input determines the output class. Decision trees are self-interpretable, meaning the tree itself describes the underlying rules for classification. The advantages of the Decision Tree Classifier include its simplicity, interpretability, ability to model nonlinear data, ability to model high dimensional data, ability to work with large datasets to produce accurate results, and ability to handle outliers during training. 
\subsection{Random Forest}
A Random Forest Classifier is a supervised predictive machine learning algorithm commonly used for classification tasks. A Random Forest consists of an ensemble of decision trees, each of which provides a ``vote", or a classification, predicting class based on a majority of votes from the decision trees. Random forests generally outperform decision trees in terms of accuracy; however, the random forest is a \textit{blackbox}, a model unable to describe its underlying rules for classification, sacrificing the interpretability of the decision tree.
\subsection{Naive Bayes}
A Naive Bayes Classifier is a supervised probabilistic machine learning algorithm commonly used for classification tasks. It relies on Bayes' Theorem, which is a theorem of conditional probabilities. Naive Bayes assumes strong independence between the input features. We use the Gaussian Naive Bayes Classifier based on the assumption that each input feature is normally distributed. The Naive Bayes Classifier is simple (and, therefore, computationally fast), scalable (requiring parameters linear in the number of features), and works well with high-dimensional data.
\subsection{Logistic Regression}
A Logistic Regression Classifier is a supervised predictive machine learning algorithm used for classification tasks. It is a type of generalized Linear Regression algorithm with a complex cost function. The cost function used here is the Sigmoid function, which continuously maps real-valued numbers between 0 and 1. We partition these continuously-mapped values from the cost function by various threshold values to determine output class. We include Logistic Regression in this work based on its speed, non-assumption of feature independence, and ubiquity in multi-class classification.
\subsection{Artificial Neural Network}
An artificial neural network is a supervised predictive machine learning technique that is successful in classification tasks in various applications \cite{b16,b17,b18}. A feed-forward neural network is a subset of neural networks in general where there are no cycles formed between neurons. We choose to use a specific type of feed-forward neural network, the multi-layer perceptron (MLP). An MLP consists of at least one hidden layer of neurons, as opposed to a single-layer perceptron, which does not have a hidden layer of neurons separate from the input and output layers. We propose to use a shallow neural network, as opposed to a deep neural network, which is a subset of the multi-layer perceptron. For a neural network to be \textit{shallow}, it means that the network has exactly one hidden layer with any number of neurons. 
\subsubsection{Architecture}
\begin{figure}
\centerline{\begin{tikzpicture}[shorten >=1pt,->,draw=black!50, node distance=\layersep]
    \tikzstyle{every pin edge}=[<-,shorten <=1pt]
    \tikzstyle{neuron}=[circle,fill=black!25,minimum size=17pt,inner sep=0pt]
    \tikzstyle{input neuron}=[neuron, fill=green!50];
    \tikzstyle{output neuron}=[neuron, fill=red!50];
    \tikzstyle{hidden neuron}=[neuron, fill=blue!50];
    \tikzstyle{annot} = [text width=4em, text centered]

    \foreach \name / \y in {1,...,4}
        \node[input neuron] (I-\name) at (0,-\y) {};

    \foreach \name / \y in {1,...,5}
        \path[yshift=0.5cm]
            node[hidden neuron] (H-\name) at (\layersep,-\y cm) {};

    \node[output neuron, right of=H-3] (O) {};

    \foreach \source in {1,...,4}
        \foreach \dest in {1,...,5}
            \path (I-\source) edge (H-\dest);

    \foreach \source in {1,...,5}
        \path (H-\source) edge (O);

    \node[annot,above of=H-1, node distance=1cm] (hl) {Hidden layer};
    \node[annot,left of=hl] {Input layer};
    \node[annot,right of=hl] {Output layer};
\end{tikzpicture}}
\caption{Fully-connected artificial neural network with shape (210,256,3), meaning the input layer has 210 neurons, the single hidden layer has 256 neurons, and the output layer has 3 neurons.}
\label{fig:arc}
\end{figure}
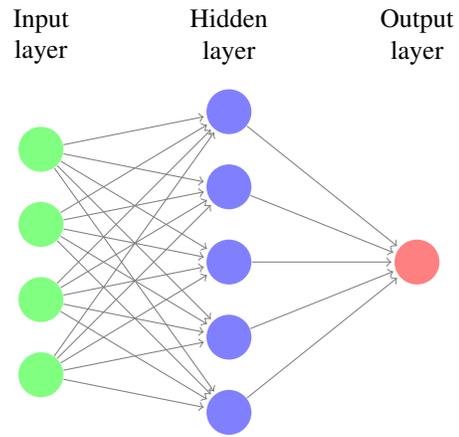

For this investigation, we propose a fully connected artificial neural network with one hidden layer. The network has a single input layer where the number of neurons equals the number of input features and a single output layer where the number of neurons equals the number of classes. Thus, we use a neural network architecture similar to Figure.\ref{fig:arc} with 210 input neurons, 256 hidden layer neurons, and three output neurons. 
Non-linear activation functions allow neural networks to learn complex patterns. We use the rectified linear unit activation function after the hidden layer to introduce this non-linearity into the model. We use the softmax activation function after the output layer. Softmax is commonly used in multi-class classification problems, more general than sigmoid, and maps output into the range between 0 and 1, making it a good function for determining class. We use the \textit{backpropagation} learning technique and the \textit{adam} optimizer \cite{adam}. 
\subsection{Evaluation of Model Performance}
We evaluate the performance of each of the classifiers by comparing the predictions of each model to the manual scoring of domain experts. Testing accuracy is a ratio of the number of correctly classified samples by the model to the total number of samples. The primary measure of performance used in this investigation will be testing accuracy.

However, testing accuracy does not always fully describe the performance of a machine learning model. Oftentimes it is important to maximize the true positive classifications or to minimize the false positive classifications. In order to measure these situations, we have the proportions called \textit{precision} and \textit{recall}. \textit{Precision} is the proportion of model classifications that correspond to the sample's true class. \textit{Recall} is the proportion of samples that are actually classified by the model correctly. In addition to these metrics, we also provide \textit{F1 Score}. The \textit{F1 Score} is the harmonic mean of precision and recall. Finally, \textit{AUC}, or Area Under the receiver operating characteristic Curve, represents the probability that the model ranks a random positive example higher than a random negative example. We choose to display testing classifications for each machine learning model in the form of a Confusion Matrix. In each confusion matrix, class 0 corresponds to class P, class 1 corresponds to class S, and class 2 corresponds to class W. Accuracy, precision, recall, F1 score, and AUC can all be derived from a confusion matrix.

\def \testingacc{0.9331281781196594}
\def \testingloss{0.21030423045158386}
\def \testingprecision{0.9390923380851746}
\def \testingrecall{0.9279104471206665}
\def \testingauc{0.9867308735847473}
\def \trainingacc{0.9478}
\def \trainingloss{0.1544}
\def \trainingprecision{0.9522}
\def \trainingrecall{0.9435}
\def \trainingauc{0.9924}
\def \computationtime{220.66}
\section{Results}
We evaluate the performance of the aforementioned methodologies primarily through testing accuracy. The testing accuracies for each methodology are described in Table.~\ref{table:acc}. F1 Score and AUC Score are provided for each machine learning technique; however, it will suffice to focus solely on accuracy.\begin{table}[!htbp]
\caption{Performance of various machine learning models in sleep stage classification.}
\centerline{\begin{tabular}{lllll}
\multirow{2}{*}{\textit{\textbf{Classifier}}} & \multirow{2}{*}{\textit{\textbf{Accuracy}}} & \multirow{2}{*}{\textit{\textbf{F1 Score}}} & \multirow{2}{*}{\textit{\textbf{AUC Score}}}\\\\
\textit{Random Forest} & 95.78\% &96\%  &.9954\\
\textit{ANN} & 93.31\% &93.9\% & .9867 \\    
\textit{Decision Tree Classifier} & 92.77\%  &93\% &.9427\\
\textit{Logistic Regression} & 77.33\%  &77\%  &.9242\\
\textit{Naive Bayes} & 74.37\%   &74\%   &.9011\\
\end{tabular}}
\label{table:acc}
\end{table}

\begin{figure}[!htbp]
    \centering
        \begin{minipage}{0.24\textwidth}
        \centerline{
        \scalebox{.23}{
        \includegraphics{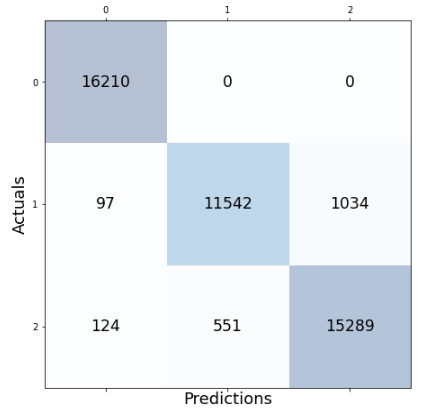}}
        }
        \caption{Confusion matrix for the Random Forest model with overall accuracy of 95.78\%.}
        \label{fig:rf_cm}
    \end{minipage}
            \begin{minipage}{0.24\textwidth}
        \centerline{
        \scalebox{.21}{
        \includegraphics{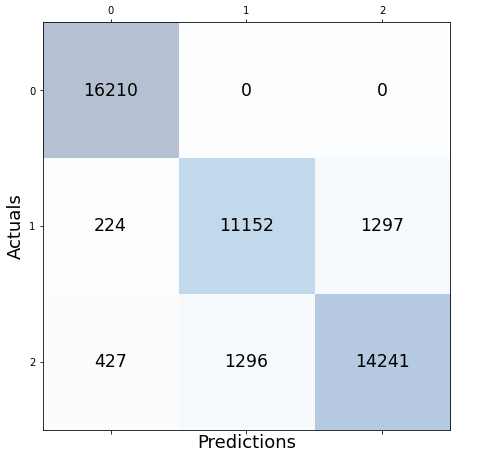}}
        }
        \caption{Confusion matrix for the Decision Tree model with overall accuracy of 92.77\%.}
        \label{fig:dt_cm}
    \end{minipage}
            \begin{minipage}{0.24\textwidth}
        \centerline{
        \scalebox{.32}{
        \includegraphics{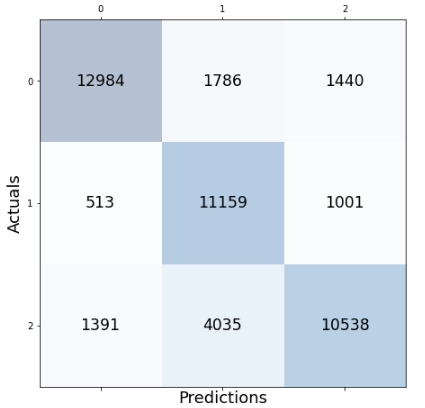}}
        }
        \caption{Confusion matrix for the Logistic Regression model with overall accuracy of 77.33\%.}
        \label{fig:lr_cm}
    \end{minipage}
            \begin{minipage}{0.24\textwidth}
        \centerline{
        \scalebox{.32}{
        \includegraphics{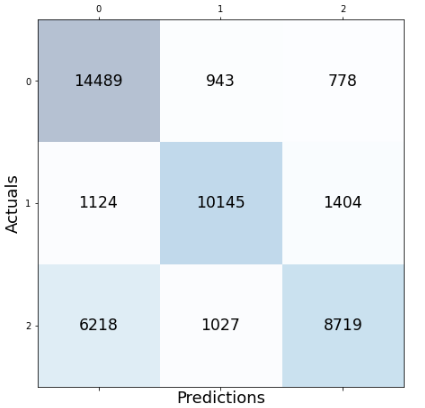}}
        }
        \caption{Confusion matrix for the Naive Bayes model with overall accuracy of 74.37\%.}
        \label{fig:nb_cm}
    \end{minipage}\hfill
\end{figure}

Notably, the highest performing classifier is the Random Forest model with a testing accuracy of 95.78\%. The confusion matrix for the testing classification of this model is shown in Fig.~\ref{fig:rf_cm}. Other metrics such as precision and recall may be derived from the confusion matrix. The model performs excellently on class 0, which corresponds to paradoxical sleep. The most frequently misclassified prediction-label pair occurred between slow-wave sleep and wakefulness. There were 1017 instances of predicting wakefulness where the true label was slow-wave and 606 samples where the model predicted slow-wave sleep where the true label was wake. 

\begin{figure}[!htbp] 
        \centerline{\scalebox{.5}{
        \import{figures}{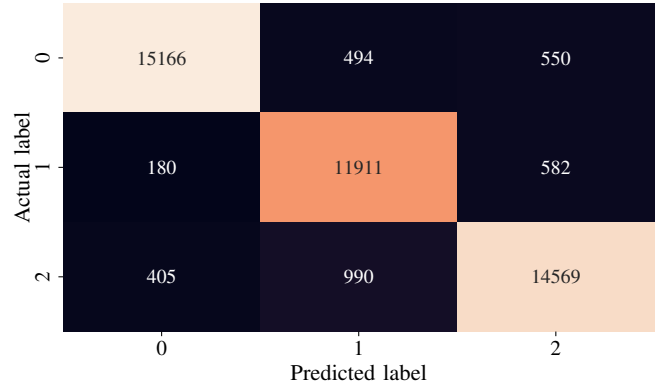}
        }}
        \caption{Confusion matrix for the Artificial Neural Network classifier with overall accuracy of 93.31\%.}
        \label{fig:ann_cm}
    \end{figure}

The second-highest performing classifier is the Artificial Neural Network with a testing accuracy of 93.31\%. Across the entire testing partition, the model achieves over 93\% categorical accuracy, which is comparable or higher than many other methodologies in the literature \cite{b5,b19,b20}. The confusion matrix for the testing classification of this model is shown in Fig.~\ref{fig:ann_cm}. The prediction-label pair that was most frequently misclassified by the ANN occurred when the model predicted slow-wave and the true label was wake. Interestingly, this is the same confusion pair that occurred most frequently in the Random Forest model. Theoretically, there is a dimension that would reliably distinguish between slow-wave and wake. Future research will investigate this relationship and attempt to distinguish more reliably and accurately between these two classes. The ANN achieved a high F1 Score and AUC Score of 93.9\% and .9867, respectively. After training for only 50 epochs, the model achieves a testing categorical accuracy of 94.01\%, testing precision of 94.54\%, testing recall of 93.43\%, and a testing AUC of 99\%. The value of these metrics over the period of training are shown in Fig.~\ref{fig:loss}, Fig.~\ref{fig:precision}, and Fig.~\ref{fig:recall}.

\begin{figure}[!htbp]
    \centering
    \begin{minipage}{0.24\textwidth}
        \centerline{\scalebox{.25}{
        \import{figures}{loss.pgf}
        }}\caption{Loss curve for training and validation data over 50 epochs.}\label{fig:loss}
    \end{minipage}\hfill
    \begin{minipage}{0.24\textwidth}
        \centerline{\scalebox{.25}{
        \import{figures}{categorical_accuracy.pgf}
        }}
        \caption{Accuracy curve for training and validation data over 50 epochs.}
        \label{fig:acc}
    \end{minipage}
        \begin{minipage}{0.24\textwidth}
        \centerline{\scalebox{.25}{
        \import{figures}{precision.pgf}
        }}
        \caption{Precision curve for training and validation data over 50 epochs.}
        \label{fig:precision}
    \end{minipage}
        \begin{minipage}{0.24\textwidth}
        \centerline{\scalebox{.25}{
        \import{figures}{recall.pgf}
        }}
        \caption{Recall curve for training and validation data over 50 epochs.}
        \label{fig:recall}
    \end{minipage}
\end{figure}
Not only did this model achieve a high degree of categorical accuracy, precision, and recall, but it did this with very little training and computation. All training instances consisted of only 50 epochs, each of which took approximately \computationtime~seconds when run on a 2.4 GHz 7th Generation Intel(R) Core(T M) i7-7700HQ Quad-Core Processor with 16GB (2x8GB) DDR4 at 2400Mhz.

The Decision Tree model performed well with an accuracy of 92.77\%; however, we do not discuss this model any further because it is a special case of the Random Forest. Logistic Regression and Naive Bayes had testing accuracies of 77.33\% and 74.37\%, respectively. These low accuracies suggest that this vigilance state classification problem is not trivial and that the accuracies reached by the Random Forest and the Artifical Neural Network are remarkable achievements.
\section{Conclusion}
Many works which rely on multiple EEG channels have successfully produced automated vigilance state classifiers; however, the hardware required to obtain such signals produces undesirable disadvantages. Here we evaluate various machine learning techniques in vigilance state classification based on a single EEG channel. Random Forests and Artificial Neural Networks produced remarkable accuracies of approximately 96\% and 93\%. In evaluation of these classification techniques against human scoring as ground truth, we note that humans may make misclassifications. In fact, due to the rigidly patterned nature of the data and the power of these statistical models, it would be appropriate to reevaluate human scores based on model classifications. Future research will have a domain expert label polysomnograms from the additional 147 hours of unscored data aided by the model from this work. Additionally, future research will have a domain expert reevaluate classification pairs that confused the models in this work. Here we achieved two accurate and reliable models that can be used immediately in an automated vigilance state classifier and may be reinforced in future work.
\bibliographystyle{IEEEtran}
\bibliography{references}
\end{document}